\documentclass[10pt,twocolumn,letterpaper]{article}

\usepackage{wacv}
\usepackage{times}
\usepackage{epsfig}
\usepackage{graphicx}
\usepackage{amsmath}
\usepackage{amssymb}
\usepackage{booktabs}
\def\wacvPaperID{1090} % Enter the WACV Paper ID here

\wacvapplicationstrack % Uncomment this line if you are submitting to the Applications Track.

\wacvfinalcopy % *** Uncomment this line for the final submission

\ifwacvfinal
\usepackage[breaklinks=true,bookmarks=false]{hyperref}
\else
\usepackage[pagebackref=true,breaklinks=true,colorlinks,bookmarks=false]{hyperref}
\fi

\graphicspath{{./}}

\usepackage{lmodern,bm}
\usepackage{soul}
\sethlcolor{red}

\usepackage{comment}
\usepackage{adjustbox}

\begin{document}

%%%%%%%%% TITLE
\title{CellTranspose: Few-shot Domain Adaptation for Cellular Instance Segmentation}

\author{Matthew R. Keaton \qquad Ram J. Zaveri \qquad Gianfranco Doretto\\
West Virginia University\\
Morgantown, WV 26506\\
{\tt\small \{mrkeaton, rz0012, gidoretto\}@mix.wvu.edu }
% \and
}

\maketitle
\thispagestyle{empty}

%%%%%%%%% ABSTRACT
\begin{abstract}
Automated cellular instance segmentation is a process utilized for accelerating biological research for the past two decades, and recent advancements have produced higher quality results with less effort from the biologist. Most current endeavors focus on completely cutting the researcher out of the picture by generating highly generalized models. However, these models invariably fail when faced with novel data, distributed differently than the ones used for training. Rather than approaching the problem with methods that presume the availability of large amounts of target data and computing power for retraining, in this work we address the even greater challenge of designing an approach that requires minimal amounts of new annotated data as well as training time.
We do so by designing specialized contrastive losses that leverage the few annotated samples very efficiently. A large set of results show that 3 to 5 annotations lead to models with accuracy that: 1) significantly mitigate the covariate shift effects; 2) matches or surpasses other adaptation methods; 3) even approaches methods that have been fully retrained on the target distribution. The adaptation training is only a few minutes, paving a path towards a balance between model performance, computing requirements and expert-level annotation needs.
\end{abstract}

%%%%%%%%% BODY TEXT

\section{Introduction}
Automating the analysis of scientific imaging data via computer vision techniques is becoming increasingly convincing as our field matures. In order to accelerate scientific discovery, neural network-based methods have recently been developed to automatically segment and count individual instances of cells in laboratory-produced imaging data~\cite{Stringer2021-yw,Weigert_2020_WACV,Greenwald2021-uj}. Such type of data acquisitions exhibit a remarkable variability, which is due to the large variety of imaging modalities being used, the different types of tissues and how they are processed.

Current approaches for the specific task of cell instance segmentation are mainly based on supervised learning. They are trained on large datasets in an attempt to compensate for the diversity of the new data they are meant to be used on. However, new data to process will  very likely not be distributed in the same way as the data used for training the models, so, they will perform the task, often with disappointing accuracy. To adress this covariate shift problem~\cite{Shimodaira2000-rp} the obvious solution is retraining the models, which is costly and time consuming because it requires manual annotation of large amounts of the new \emph{target} data we are seeking to automatically process. An alternative is to use domain adaptation methods, which attempt to adapt the model to the target data distribution. Current domain adaptation methods for segmentation are by and large tailored to imaging modalities, or specific tasks or applications that are very different than cell instance segmentation~\cite{Guan2022-qj,Xun2021-ne}. Some promising work has addressed the problem in an unsupervised manner~\cite{Liu2020-qh,Liu2021-nw}. But these approaches assume that a large fraction and amount of target data is available to undergo a relatively intense training to adapt the model.

In this work, we advocate for a more practical and scalable solution to address the need to generalize well out of distribution. We assume that a model for segmenting instances, such as cell bodies, membranes, or nuclei, has already being trained on a \emph{source} dataset. Then, by annotating only a handful of samples of the \emph{target} dataset we \emph{adapt} the model, with a low training budget, to generalize well on the new distribution. We introduce \textsl{CellTranspose}, a new approach that implements the paradigm just described, for the few-shot supervised adaption of cell instance segmentation. The approach builds on a state-of-the-art model, and we introduce new losses and a training procedure for the quick adaptation of the model to new data. 

Our framework allows for a broad range of data to be properly segmented beyond the capabilities of current generalist approaches. We show that only a small number of annotations on the target dataset are required for the model to learn to produce high-fidelity segmentations and demonstrate this both on 2-D and 3-D data. In particular, few annotated samples are sufficient also to reach adaptation levels comparable to the unsupervised adaptation models. Additionally, CellTranspose affords a much faster training scheme as compared to training a model with similar accuracy from scratch.

%%% Local Variables:
%%% mode: latex
%%% TeX-master: "main"
%%% End:

%%%%%%% KEEP A LOT OF COMMENTED PARAGRAPHS FOR JOURNAL %%%%%%%
\section{Related Work}

\subsection{Cellular Instance Segmentation}

The current top approach for cell instance segmentation is Cellpose \cite{Stringer2021-yw}.
Like many recent segmentation algorithms, Cellpose’s underlying model is a variant of U-Net \cite{Ronneberger2015-io}, outputing a pixel-wise mask prediction.
By itself, this only produces a semantic segmentation: the output mask merely determines the class of each pixel (foreground/cell or background), making it impossible to delineate between individual cells when they are clustered together. Deriving inspiration from the traditional watershed algorithm \cite{Beucher1979-up} and a gradient-based deep pose recognition algorithm called OpenPose \cite{8765346}, Cellpose contains two additional outputs corresponding to predicted gradients towards each pixel’s associated cell source location, one in the $x$-direction and one in the $y$-direction. Through an iterative process, the gradient of each pixel is followed to neighboring pixels until a ``source'' pixel representing a cell center is encountered. Each pixel directly linked to the source is then considered a part of the cell instance, in this way constructing segmentations of individual cells.

Other cellular segmentation methods have been proposed recently. Similar to Cellpose, Mesmer \cite{Greenwald2021-uj} represents another variant of ``deep watershed'' algorithms, also producing a pixel-wise mask prediction, but instead of the two flow outputs it generates an ``inner distance transform'' to predict the distance of each pixel to a cell's center.
StarDist \cite{schmidt2018} and StarDist-3D \cite{Weigert_2020_WACV} predict object centers and then approximate the distance from the given location to cell boundaries at fixed angles from the center. The points defined by these predictions are then connected to each neighboring point to produce the outline of the cell’s mask. Other approaches include NuSeT  \cite{Yang2020-yn}, which also builds on the watershed algorithm, and DenoiSeg \cite{Buchholz2020-su}, which utilizes a joint learning strategy in concert with a denoising task to produce better accuracy on noisy samples.

\subsection{Domain Adaptation for Medical Imaging}

A common issue in deep learning practice is the scarcity of labeled data for many tasks and the often unrealistic requirement to produce vast amounts of costly annotations in order to appropriately train a model. Although many efforts have been made to produce ``generalist'' models which are invariant across different datasets for a specific task, doing so is intractable in real-world applications where data has intrinsically high variability. Domain adaptation allows for taking a high-performing model trained on a large dataset and adapting it to work on new data representing some target domain. This enables the model to take advantage of learned low-level features acquired from the larger source dataset while tuning to the specific features of the target dataset, requiring less target data.

Due to this lower dependence on manual annotation from experts, a vast amount of research on domain adaptation-based approaches has been unleashed on the medical imaging field~\cite{Guan2022-qj,Xun2021-ne}. In addition to a few domain adaptation approaches on semantic segmentation for various imaging modalities including brain tumors~\cite{Ghafoorian2017-kz,Van_Opbroek2015-wq,Goetz2016-az}, whole tissue~\cite{Zhu2020-mv}, and organelles~\cite{Bermudez-Chacon2018-bf}, several unsupervised mechanisms for semantic segmentation~\cite{Yan2019-cv,Kamnitsas2017-uh,Javanmardi2018-ym,Wang2019-ue,Panfilov2019-gc,Dou2019-ey,Bateson2019-of,Gholami2019-li,Jue2019-co,Jiang2018-vm,Zhang2018-yj,Zhang2020-eu,Chen2019-ov,Yan2019-wf,Yang2019-cq,Perone2019-wy,Shanis2019-cz,Orbes-Arteaga2019-pu,Karani2018-ln} and the more challenging instance segmentation task~\cite{Liu2020-qh,Liu2021-nw,Yang2021-qs,Li2021-mg} have been proposed. Generally, these unsupervised methods aim to learn representations of a given target domain without the use of any annotations. This greatly reduces the need for effort from medical experts while producing better segmentations than non-adapted models. However, this is accompanied by a clear trade-off with model accuracy, and the training scheme, in addition to more complex model architectures, causing training time to increase heavily. There is also an intrinsic expectation for the shape and relative size of segmentations from the target data to match closely with those of the source dataset, constraining such approaches to nuclei segmentation rather than more heterogeneous full-cell segmentations. These methods are also notably not ``generalist,'' and tend to focus on and perform best on a particular cell or imaging type.

\subsection{Few-shot Domain Adaptation}

In order to accommodate many real-world settings, one set of techniques known as ``few-shot'' domain adaptation aims to balance the demand for annotations with overall model performance. Since at least one large dataset is often available for a given task, many of the weaknesses posed by few-shot learning on its own can be avoided. Some popular techniques for addressing domain adaptation with a limited number of target samples includes utilizing adversarial techniques to increase the confusion between the source and target domain \cite{Motiian2017-sr}, using meta-learning approaches to use tasks from one domain for adversarially learning tasks in a smaller, novel domain \cite{Sahoo2018-ji}, and generating prototypes based on learned embeddings from the few target samples \cite{Zhao2021-tt}, a technique common to few-shot learning.

One which has proven effective across multiple paradigms is the use of contrastive losses \cite{Hadsell2006-it}. A contrastive loss function encourages each sample from the target domain to be represented similarly to one or more samples of the same class (positive samples) in the source domain while simultaneously pushing its representation away from samples from other classes (negative samples) within the source domain. This approach has proven successful in a number of scenarios including domain adaptation \cite{Li2021-pu,wang2020DenseCL,Liu2021-li}, unsupervised learning \cite{Wang2020-ld,Zhuang2019-lx,Melas-Kyriazi2021-pt,Van_den_Oord2018-vn,He2020-kk,Caron2020-xw}, self-supervised learning \cite{Chen2020-zr,Yeh2021-cf,Misra2020-fh}, meta-learning \cite{Sahoo2018-ji}, and even few-shot domain adaptation \cite{Motiian2017-ow}.

%%% Local Variables:
%%% mode: latex
%%% TeX-master: "main"
%%% End:

\section{Problem Definition}

\begin{figure}[t!]
\centering
\includegraphics[width=0.48\textwidth]{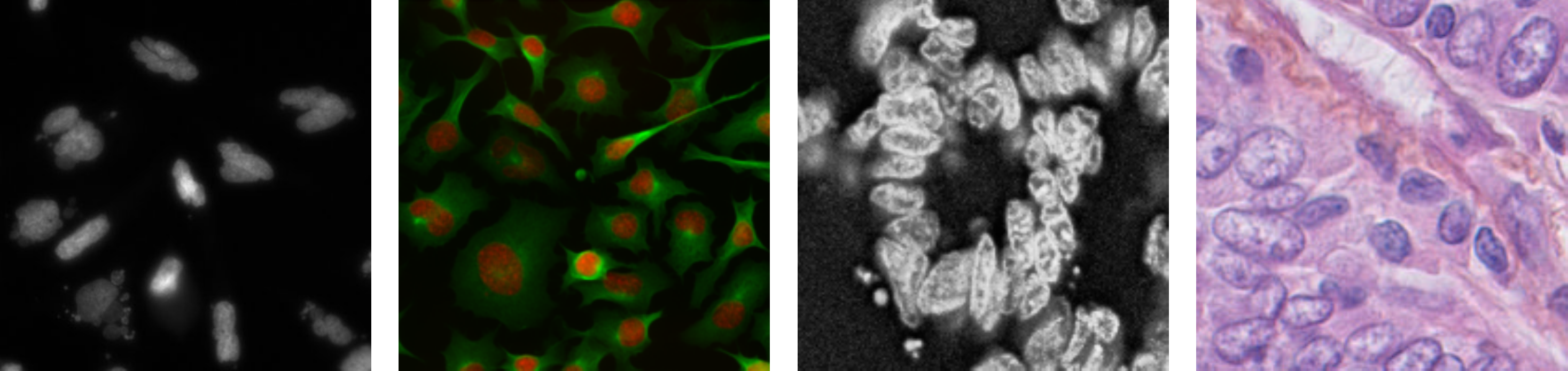}
\caption{\textbf{Variability of microscopy data.} Image samples highlighting the variability of cell images. From left to right: Human U20S cells with Hoechst and phalloidin stains from BBBC006~\cite{Ljosa2012-si}; Neuroblastoma cells labelled with phalloidin and DAPI stains from the Cell Image Library~\cite{cil_dataset}; GI tissue cells imaged by co-detection by indexing (CODEX) from Tissuenet~\cite{Greenwald2021-uj}; Breast cancer cells, using hematoxylin and eosin (H\&E) stain from TNBC~\cite{Naylor2019-ak}.}
\label{fig:modality_samples}
\vspace{-5mm}
\end{figure}

We are interested in the segmentation of instances such as cell bodies, membranes, or nuclei which are present in cell images. Such image structures have a remarkable variability because they are acquired with a variety of microscopy techniques,
%/fluorescent markers
come from a diverse range of tissues, and present very different shapes. See Figure~\ref{fig:modality_samples}. Therefore, even state-of-the-art supervised instance-based segmentation approaches~\cite{Stringer2021-yw,Greenwald2021-uj}, which claim to offer a \emph{generalist} solution, unfortunately experience rapid performance degradation when facing the challenge of data distribution shift~\cite{Shimodaira2000-rp}. In essence, when the so-called generalist model is tasked to operate on data distributed differently than the data it was trained on, expectations are unmet.

Given how frequent the need to generalize out-of-distribution is for the task at hand, and given how costly and time consuming, if not completely impractical, the process of collecting and annotating sufficient target data for retraining the model can be, we reframe the problem as one of \emph{few-shot} learning. The intent is to significantly widen the range of image variability handled by current generalist solutions. Specifically, given a \emph{source} dataset $\mathcal{D}^s$ made of labeled images with which a generalist model could be trained, the learning task is to \emph{adapt} such a model to generalize well on a \emph{target} dataset $\mathcal{D}^t$, which is distributed differently than $\mathcal{D}^s$. To perform such adaptation, the user is required to label only a \emph{minimal} amount of target data, let us say $K$ data instances for a $K$-shot learning. In Section~\ref{sec-adaptation} we clarify the meaning of a single shot.

Note that simple fine-tuning of a generalist model on very few target samples is not a feasible solution due to the obvious overfitting challenges~\cite{Goodfellow2016-iu}, which is why we introduce a method that aims at a very low budget in terms of the amount of target data to annotate, as well as the time for adaptation training. Given the specific setting of our problem, to the best of our knowledge, this is the first work to propose a method for few-shot supervised domain adaptation for the task of cell instance segmentation.
\begin{figure*}[t!]
\centering
\includegraphics[height=4.6cm]{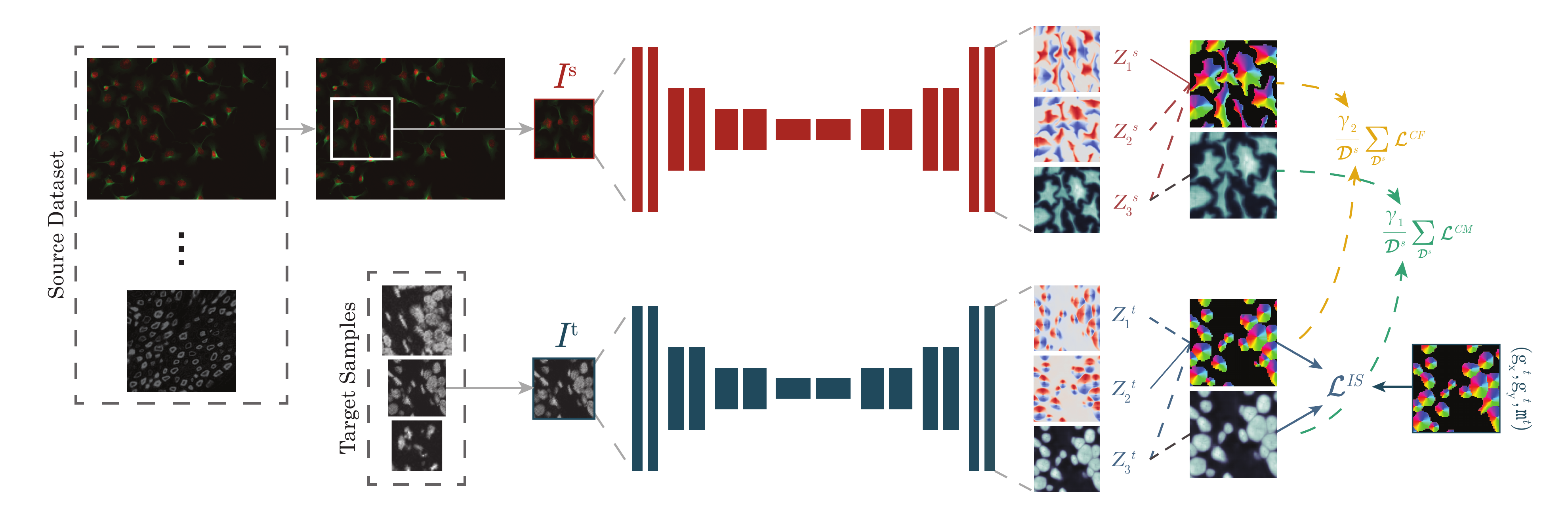}
\caption{\textbf{Architecture approach.} Comprehensive illustration of our contrastive learning-based few-shot cellular instance segmentation approach.}
\label{fig:adaptation_approach}
\vspace{-5mm}
\end{figure*}

\section{Approach}

Our approach, shown in Figure~\ref{fig:adaptation_approach}, assumes that some large labeled source dataset $\mathcal{D}^s$ can be used to initially train the model. However, regardless of the variety of cell images within $\mathcal{D}^s$, there will always conceivably be new data samples outside of the source distribution. This is due to limitations in both model and dataset size. We also assume that given some out-of-distribution set of target samples, annotation is costly, and therefore only a few samples will be labeled. In the following subsections, we first describe the pretrained method we rely on, and then introduce the proposed few-shot supervised adaptation model.

\subsection{Pretrained Model}
\label{sec-pretrained-model}

To provide a robust starting point, we use the state-of-the-art cell instance segmentation approach Cellpose~\cite{Stringer2021-yw}, which we summarize here to introduce notation and make the paper self-contained. Given an image $I$, the model uses a network $f$ to produce a dense, pixel-wise feature $\bm{Z} = f(I)$, where $ \bm{Z} = [Z_1, Z_2, Z_3] \in \mathbb{R}^{h \times w \times 3}$. Given the feature $\mathbf{z} = [z_1, z_2, z_3] \in \bm{Z}$ for some pixel $i$, then $\bm{z} \doteq (z_1, z_2)$ has the meaning of gradient pointing towards the center of the cell structure to which pixel $i$ belongs. Collectively, $(Z_1,Z_2)$ form a \emph{gradient flow}. Instead, $z \doteq z_3$ represents the unnormalized score indicating the probability of pixel $i$ to belong to a cell structure. Note that with this notation, the feature $\mathbf{z}$ can be written as $\mathbf{z} = [\bm{z}, z]$. The network $f$ is trained in a supervised manner with pixel-wise instance segmentation loss
\begin{equation}
  \mathcal{L}_i^{IS} = (z_1 - \mathtt{g_x})^2 + (z_2 - \mathtt{g_y})^2 + \nu H(\mathtt{m},\sigma(z))  \; ,
  \label{eq-instance-segmentation}
\end{equation}
where, for pixel $i$,  $(\mathtt{g_x}, \mathtt{g_y})$ represents the ground-truth gradient label with unit $\ell_2$-norm, $\mathtt{m} \in \{0, 1\}$ is the binary mask label indicating absence/presence of a cell structure, $\sigma(z) \doteq 1/(1+\exp(-z))$, $H$ represents the binary cross-entropy, and $\nu$ is a hyperparameter set to $0.04$. The pixel-wise loss contributions are then aggregated into a final loss $\mathcal{L}^{IS} = \sum \mathcal{L}_i^{IS}$ for the image $I$.

Given the feature $\bm{Z}$, the cell instance segmentation head $g$ produces the mask $Y = g(\bm{Z})$, where for pixel $i$, the predicted label $y$ is a number in the set $\{0, 1, \cdots, N \}$, with $N$ being the total number of cell instances segmented. Pixels belonging to the same cell instance have same label, and $y=0$ indicates absence of a cell instance. For details on how $g$ is implemented the reader can consult~\cite{Stringer2021-yw}. See also Figure~\ref{fig:adaptation_approach}.

\subsection{Adaptation Model}

Consider now a target image $I \in \mathcal{D}^t$. If $I$ was drawn from the same distribution from which source images composing $\mathcal{D}^s$ were drawn, we would expect the following to be true: For a target pixel $i$ with label $(\mathtt{g_x}^t, \mathtt{g_y}^t, \mathtt{m}^t)$, its feature $\mathbf{z}^t$ should be very close to the features of pixels in the source dataset that have the same label. However, in presence of a \emph{domain shift} this is generally untrue, which leads to performance deterioration of the cell instance segmentation process, when we use the feature network $f$ on $I$, followed by the instance segmentation head $g$. Therefore, provided that the labels of some target image pixels are available, we design a method for reversing the effects of domain shift, by adapting $f$ to generalize well on the target dataset $\mathcal{D}^t$. Given the distinct predictive nature (continuous vs. discrete) of the gradient flow features $\bm{ z} = (z_1, z_2)$, from the mask feature $z$, we proceed by designing adaptation losses for each case.

\subsubsection{Contrastive Flow Loss}

In order to align $\bm{ z}^t$ with the gradient flow features of source pixels with same label, we set up a contrastive prediction task~\cite{Chen2020-zr}. We identify a \emph{positive} source pixel with binary label $\mathtt{m}^s_+ = \mathtt{m}^t = 1$, and with gradient flow features $\bm{ z}^s_+$ that best matches the label $(\mathtt{g_x}^t, \mathtt{g_y}^t)$, according to a similarity measure, like cosine similarity $s(\bm{u}, \bm{v}) \doteq \bm{u}^{\top} \bm{v} / \|\bm{u} \| \| \bm{v} \|$, where $\| \cdot \|$ denotes $\ell_2$-norm. Then, we compose a set of \emph{negative} source gradient flow features $\mathcal{N}_i = \{ \bm{ z}^s_- \; | \; s(\bm{ z}^s_+, \bm{ z}^s_-) < \delta, \mathtt{m}^s_- = 1 \}$, where $\delta$ is a suitable constant hyperparameter threshold that we choose. Now we can use a contrastive loss function for pixel $i$ that attempts to pull the positive pair $(\bm{ z}^t, \bm{ z}^s_+)$ closer, while pushing apart every negative pair $(\bm{ z}^t, \bm{ z}^s_-)$, for $\bm{ z}^s_- \in \mathcal{N}_i$. The loss function we used, which we name \emph{contrastive flow loss}, is
\begin{equation}
  \small
  \mathcal{L}_i^{CF} = - \log \frac{\exp( s (\bm{ z}^t, \bm{ z}^s_+)/\tau )}{ \exp( s (\bm{ z}^t, \bm{ z}^s_+)/\tau ) + \sum_{\bm{ z}^s_- \in \mathcal{N}_i} \exp( s (\bm{ z}^t, \bm{ z}^s_-)/\tau ) }
  \label{eq-contrastive-flow-loss}
\end{equation}
where $\tau$ denotes a temperature parameter. Note that~\eqref{eq-contrastive-flow-loss} addresses only the directional alignment of the gradient flow features. This is also what matters most, because the segmentation head $g$ uses only that information for assigning pixels to instances, not the gradient magnitude. However, in Section~\ref{sec-adaptation} we will see that~\eqref{eq-instance-segmentation} is also applied on target data, which does encourage gradients to have unit magnitude.

An important component of the loss is how we mine the negative source features. Specifically, $\mathcal{N}_i$ is composed by selecting the features closest to the positive source feature $\bm{ z}^s_+$. This hard-mining strategy removes the need to consider large amounts of negative source features, because it makes them less informative. This leads to faster training, and usually more rapid convergence and better model performance, as it was shown also in~\cite{Wang2020-ld}. Moreover, for a given target pixel $i$ we extract one positive feature, and all the $|\mathcal{N}_i|$ negative features from one source image. We set $|\mathcal{N}_i|$ to be the same for every target pixel, and generally, there is a sufficient number of negative features such that the similarity between $\bm{ z}^s_+$ and each negative feature is approximately equal to $\delta$. Therefore, the $\mathcal{N}_i$ features are roughly equally divided between those at $\cos ^{-1} \delta$ radians clockwise and those at $\cos ^{-1} \delta$ radians counterclockwise from $\bm{ z}^s_+$. Thus,
% as shown in Figure~\ref{fig:CFL},
when $\bm{ z}^t$ is off, the negative samples nearer to $\bm{ z}^t$ will produce a greater gradient than those further away.

Given a target image patch and a source image patch in the minibatch, for this pair we compute a loss contribution by aggregating all the components coming from the target pixels that belong to cell structures, so they have label $\mathtt{m}^t=1$. Let us indicate with $\mathcal{M}$ this set of pixels, then the loss for the given pair will be
\begin{equation}
\mathcal{L}^{CF} = \frac{1}{| \mathcal{M} |} \sum_{\mathcal{M}} \mathcal{L}^{CF}_i \; .
\end{equation}

\subsubsection{Contrastive Mask Loss}

The contrastive flow loss~\eqref{eq-contrastive-flow-loss} is a specialized version of a loss used in~\cite{Sohn2016-ar,Van_den_Oord2018-qw,Chen2020-zr,Wang2020-ld}, and when $\tau \rightarrow + \infty$ it converges to 
\begin{equation}
  \lim_{\tau \rightarrow + \infty} \mathcal{L}_i^{CF} = - s( \bm{ z}^t, \bm{ z}^s_+) + \lambda \sum_{\bm{ z}^s_- \in \mathcal{N}_i}  s (\bm{ z}^t, \bm{ z}^s_-) \; ,
  \label{eq-cl-loss-limit}
\end{equation}
where $\lambda$ is a hyperparameter. The loss~\eqref{eq-cl-loss-limit} was originally proposed in~\cite{Hadsell2006-er}, and a specialized form of it was introduced in~\cite{Motiian2017-ow} for supervised domain adaptation and generalization for multi-class visual classification.

Since we are interested in aligning the unnormalized binary classification score $z^t$ with the scores of the source pixels with same label, we derive a contrastive loss for segmentation, inspired by~\eqref{eq-cl-loss-limit} and~\cite{Motiian2017-ow}. This is also motivated by the fact that losses like~\eqref{eq-contrastive-flow-loss} have been proven effective with cosine similarity, which we cannot use for this task and we need to replace.

When~\eqref{eq-cl-loss-limit} is optimized, the first term aims at maximizing similarity. In our case this would be the similarity between $z^t$ and the unnormalized score of the pixel of a source image $z_+^s$ with label $\mathtt{m}^s = \mathtt{m}^t$. We can replace that term with an alignment term based on a distance, leading to a squared loss
\begin{equation}
  d(z^t,z_+^s) = \frac{1}{2} (z^t - z_+^s)^2  \; .
  \label{eq-ml-align}
\end{equation}
The second term of~\eqref{eq-cl-loss-limit}, instead, aims at minimizing similarity. In our case this would be the similarity between $z^t$ and the unnormalized score of pixels in source images with label different than $\mathtt{m}^t$. This would lead to separation between scores with opposite labels. Given one such pixel with score $z_-^s$, we can measure the similarity with $z^t$ with 
\begin{equation}
  k(z^t,z_-^s) = \frac{1}{2} \max(0,m - | z^t - z_-^s | )^2  \; ,
  \label{eq-ml-sep}
\end{equation}
where $m$ is a margin, and the loss provides gradient contributions when $z^t$ is within the margin $m$ of $z_-^s$.

\begin{figure}
\centering
\includegraphics[height=3.5cm]{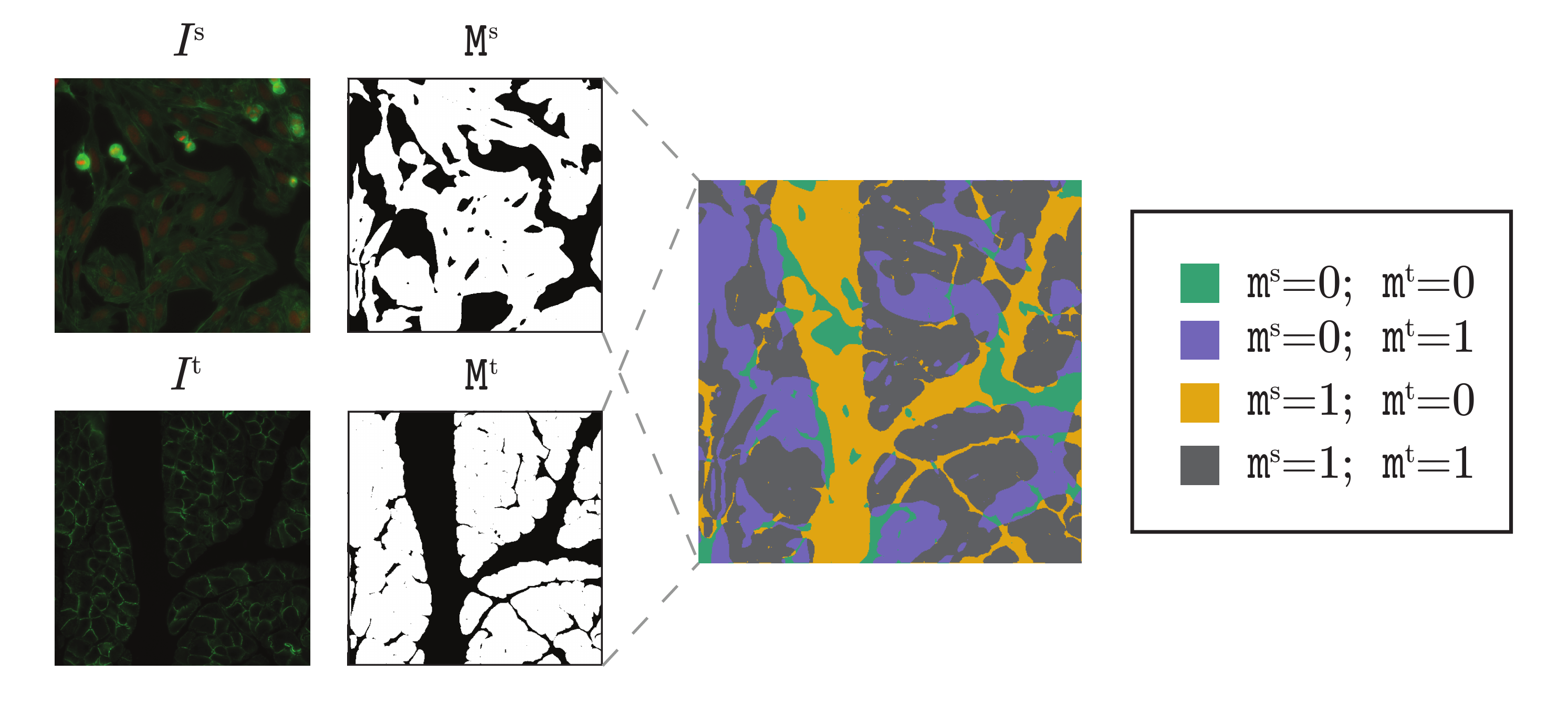}
\caption{\textbf{Contrastive Mask Loss.} {Representation of the Contrastive Mask Loss for a given source-target sample patch pair. The mask of a source and target sample are overlapped to compare representations on a per-pixel basis. Pixels belonging to the same class are encouraged to be represented similarly while pixels from different classes have their representations separated.}}
\label{fig:SAML}
\vspace{-5mm}
\end{figure}

Given a target image patch and a source image patch in a minibatch, we aggregate the losses~\eqref{eq-ml-align} and~\eqref{eq-ml-sep} as follows. Let $\mathcal{P}$ be the set of positive pairs of scores $(z^t,z_+^s)$ corresponding to pixels in the same relative positions in the target and source patches, and with same label, i.e., $\mathtt{m}^s = \mathtt{m}^t$; and let $\mathcal{N}$ be the set of negative pairs of scores $(z^t,z_-^s)$ corresponding to pixels in the same relative positions in the target and source patches but with different label, i.e., $\mathtt{m}^s \ne \mathtt{m}^t$. Then, the loss for a pair of target and source patches is
\begin{equation}
  \mathcal{L}^{CM} = \frac{1}{|\mathcal{P}|} \sum_{\mathcal{P}}   d(z^t,z_+^s) +\lambda \frac{1}{|\mathcal{N}|} \sum_{\mathcal{N}}   k(z^t,z_-^s) \; ,
  \label{eq-mask-loss}
\end{equation}
and we refer to this as the \emph{contrastive mask loss}. Note that each term of the loss is normalized with respect to the area covered by the positive and negative pairs, respectively. Also, the formation of the sets $\mathcal{P}$ and $\mathcal{N}$ is based on the comparison of the mask labels of the target and source patches, as in Figure~\ref{fig:SAML}, and the pairs originate from pixels with same relative position. This is not a strict requirement, but it is convenient because it allows forming sufficiently large sets $\mathcal{P}$ and $\mathcal{N}$, while implementing~\eqref{eq-mask-loss} is easier and faster since it takes advantage of the parallel architecture of GPUs.

\subsubsection{Few-shot Adaptation} \label{sec-adaptation}
\begin{figure*}[th!]
\centering
\includegraphics[height=5.4cm]{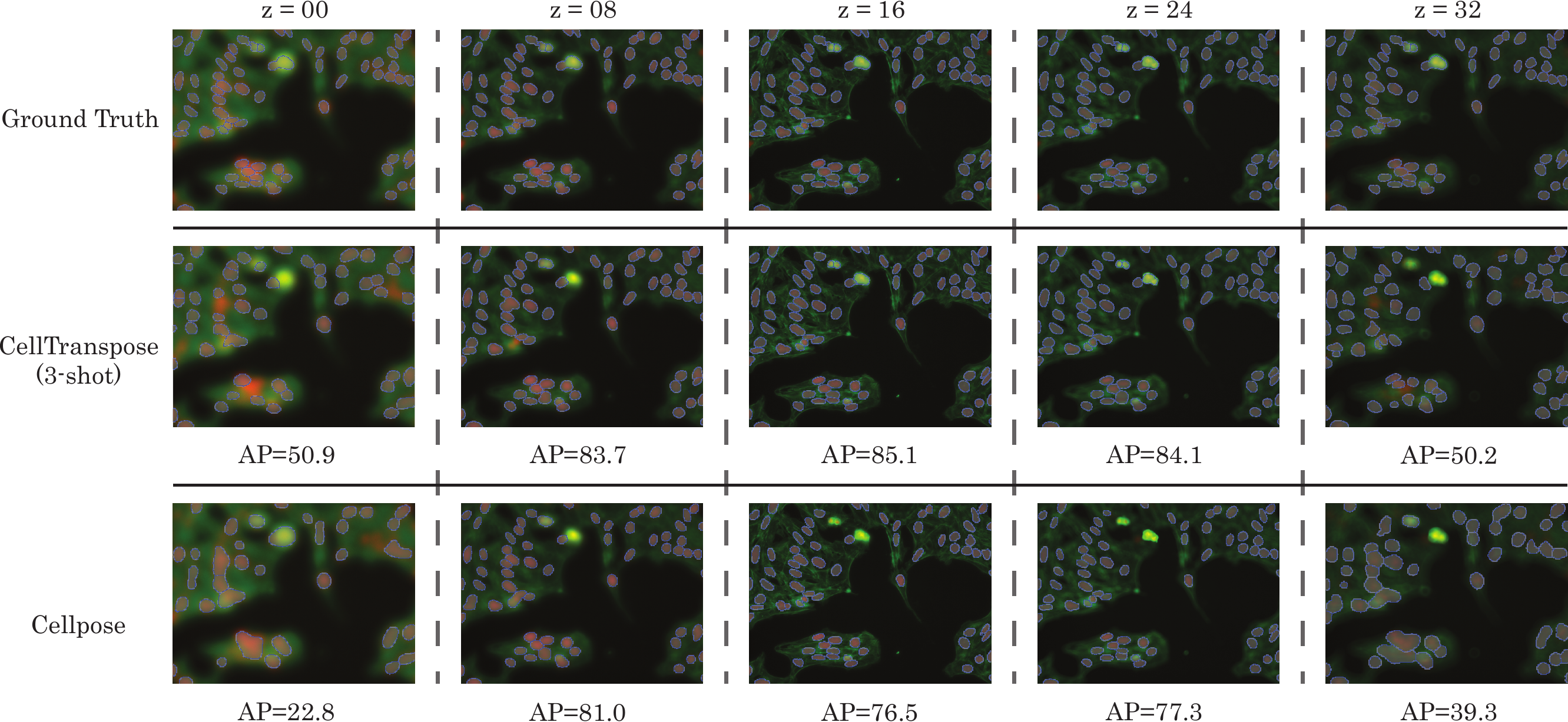}
\caption{\textbf{BBBC006 dataset.} Qualitative results on an image from the BBBC006 dataset at different levels of focus. $\mathtt{z}$=16 is considered in-focus, while values further from this are more out of focus in either direction. Segmentations are shown in gray with cell borders shown in blue. AP values are computed at 0.5 IoU.}
\label{fig:qualitative_results}
\vspace{-4mm}
\end{figure*}

Assuming that $\mathcal{D}_K^t$ is the target dataset portion of $\mathcal{D}^t$ with $K$ labeled samples, the $K$\emph{-shot adaptation learning} aims at minimizing the loss
\begin{equation}
  \mathcal{L}^{ISA} = \sum_{\mathcal{D}_K^t} \left( \mathcal{L}^{IS} + \frac{\gamma_1}{|\mathcal{D}^s|} \sum_{\mathcal{D}^s} \mathcal{L}^{CM} + \frac{\gamma_2}{|\mathcal{D}^s|} \sum_{\mathcal{D}^s} \mathcal{L}^{CF} \right) \; .
  \label{eq-adaptation}
\end{equation}
The training assumes that the generalist model has already been pretrained. The Cellpose model we used was pretrained for 500 epochs~\cite{Stringer2021-yw}. The adaptation with~\eqref{eq-adaptation} lasts 5 epochs where source images are continuously randomly paired with one of the $K$ target samples without replacement. Pairing the target samples with $1/K$-th of $\mathcal{D}^s$ in every epoch ensures that even a 1-shot adaptation can operate a significant pull of the model towards the target distribution. The adapted model is then fine-tuned on $\mathcal{D}_K^t$ with $\mathcal{L}^{IS}$ only for 5 more epochs at a very low constant learning rate.

Since this is the first attempt at few-shot domain adaptation for cellular instance segmentation, we need to define what constitutes one ``shot''. A target exemplar cell is selected manually such that its size and the relative density of cells nearby is representative of the average scenario across the target dataset. If $K>1$ cells are selected, they should better capture the variability of the dataset, and will lead to the $K$-shot scenario. For a given cell sample the cell is measured, giving the value $m_c$ in pixel units. The one shot patch is center-cropped around the cell and has size $\beta_{max} m_c w/ m_n$, where $m_n$ is the nominal cell size, $w$ is the size of the patch passed into the network, and $\beta_{max}$ is the largest scaling factor admissible. Data augmentation including translation and random cropping with factor $0.75$ to $\beta_{max}=1.25$ is part of the training to promote invariance to cell size.

%%% Local Variables:
%%% mode: latex
%%% TeX-master: "main"
%%% End:

\section{Experiments}

We evaluate the proposed approach, which we name \emph{CellTranspose}, with several datasets and multiple settings.
In absence of additional directions, the following setup is used for each experiment. The model in Section~\ref{sec-pretrained-model} is pretrained with the ``generalized'' dataset of~\cite{Stringer2021-yw} as source data $\mathcal{D}^s$. Adaptation is done as described in Section~\ref{sec-adaptation}. In general, we follow the data splitting guidelines laid out by previous works to ensure fair comparisons. In order to fit our approach to these guidelines, we draw our $K$ sample patches, to produce $\mathcal{D}^t_K$, from the training split of the dataset of the target distribution. Instead, the testing split of the target distribution dataset is used as our remaining, unlabeled portion of the target dataset $\mathcal{D}^t$.
We use SGD with initial learning rate $10^{-2}$, momentum 0.9, weight decay $10^{-5}$, and batch size of 2 provides optimal results. For the first five epochs, the learning rate decreases by a factor of 10 each epoch, and is kept constant for the remaining five. We take square patches with side length $h=w=112$, use a minimum overlap of $84$ during evaluation, and enforce a nominal cell size $m_n=30$. Additional hyperparameters are set as: $| \mathcal{N}_i |=20$, $\tau=0.1$, $m=10$, $\lambda=1$, $\gamma_1=0.05$, $\gamma_2=2$, and $\delta=0.05$. Because the source dataset used is always significantly larger than the target data, adaptation takes roughly the same amount of time regardless of the size of $K$ or the target data to which the model is adapted. Using a singular NVIDIA TITAN Xp GPU, adaptation takes approximately 5 minutes to complete for each experiment.
\begin{figure*}[th!]
\centering
\includegraphics[height=2.55cm]{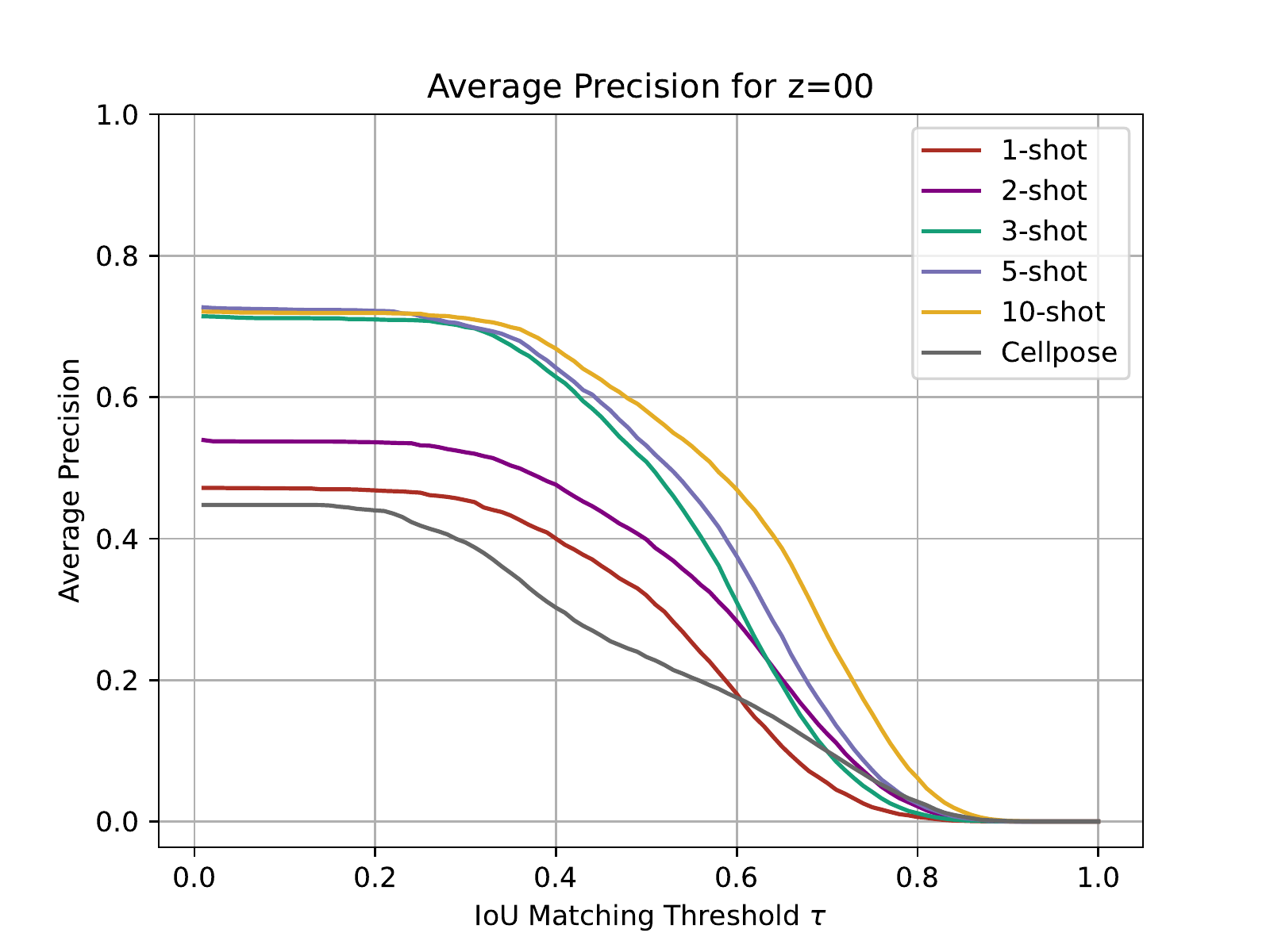}
\includegraphics[height=2.55cm]{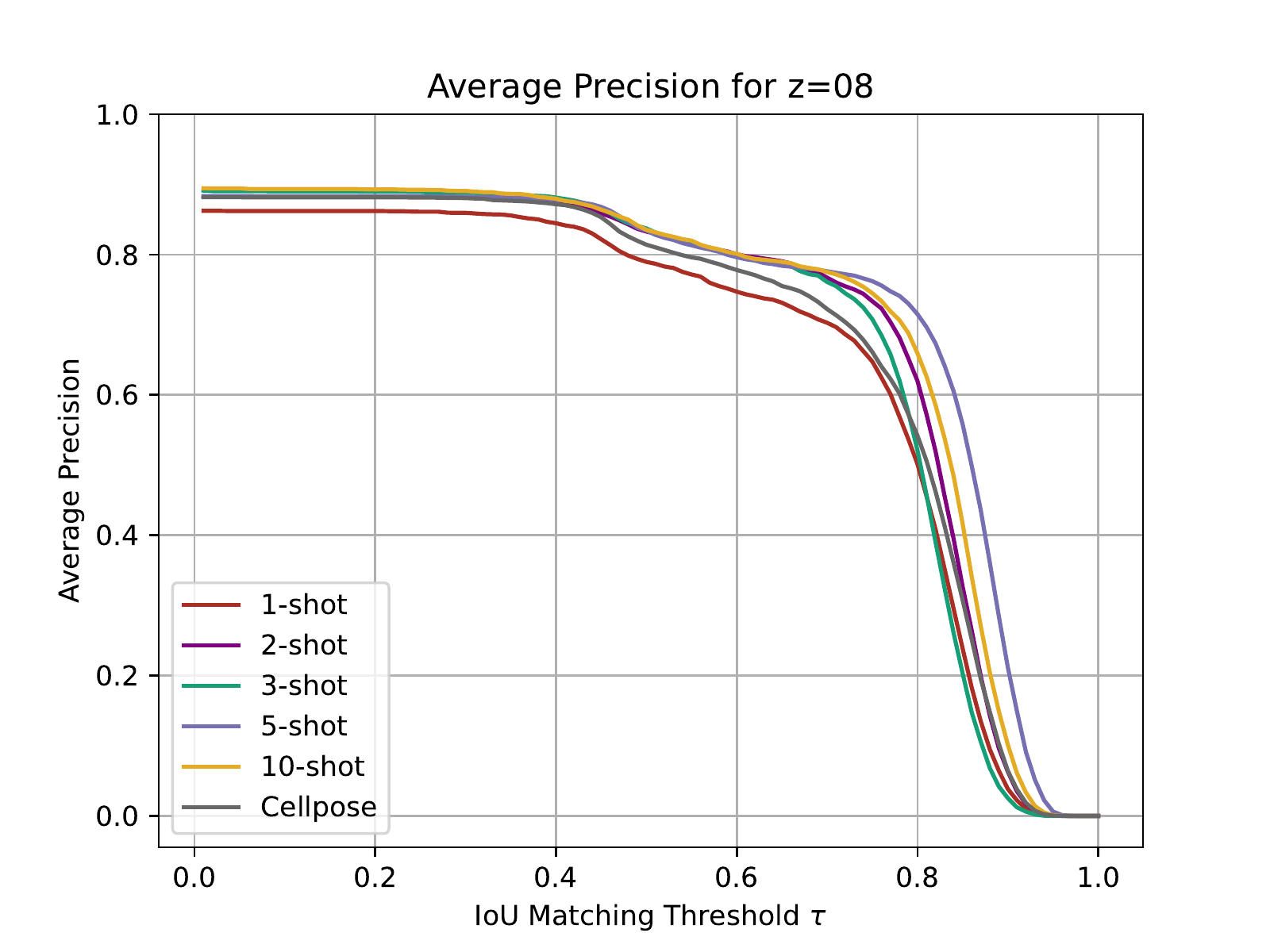}
\includegraphics[height=2.55cm]{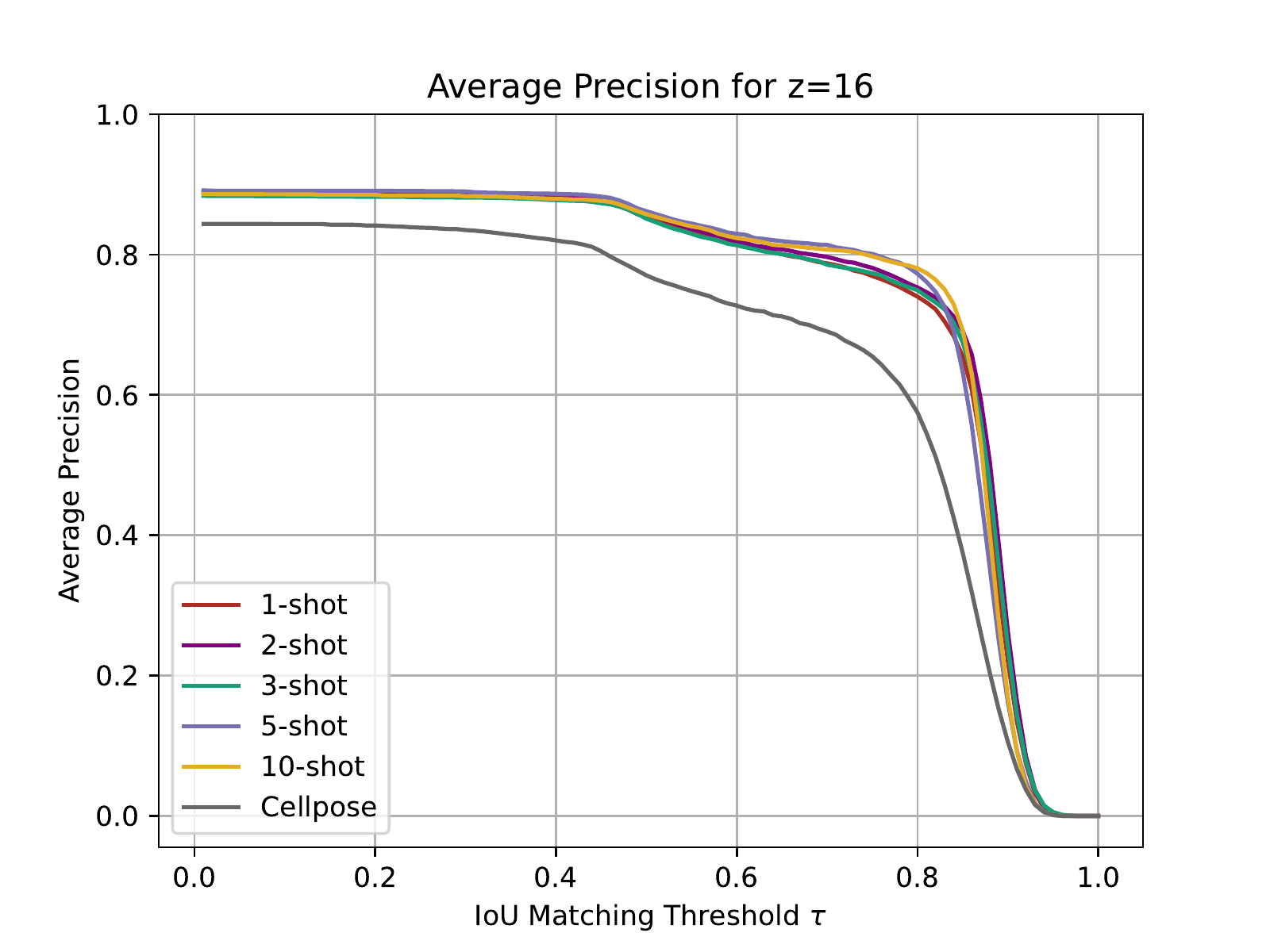}
\includegraphics[height=2.55cm]{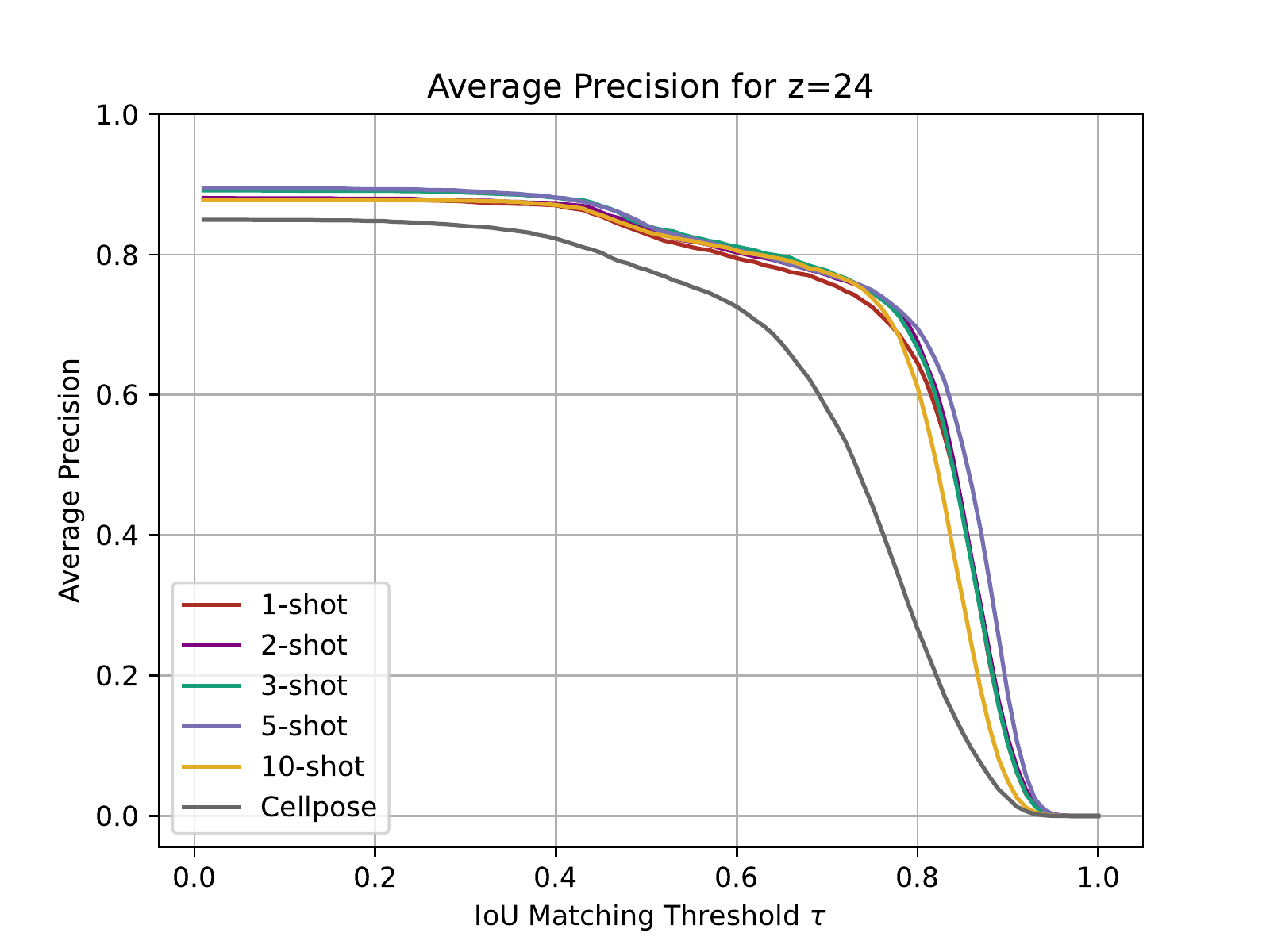}
\includegraphics[height=2.55cm]{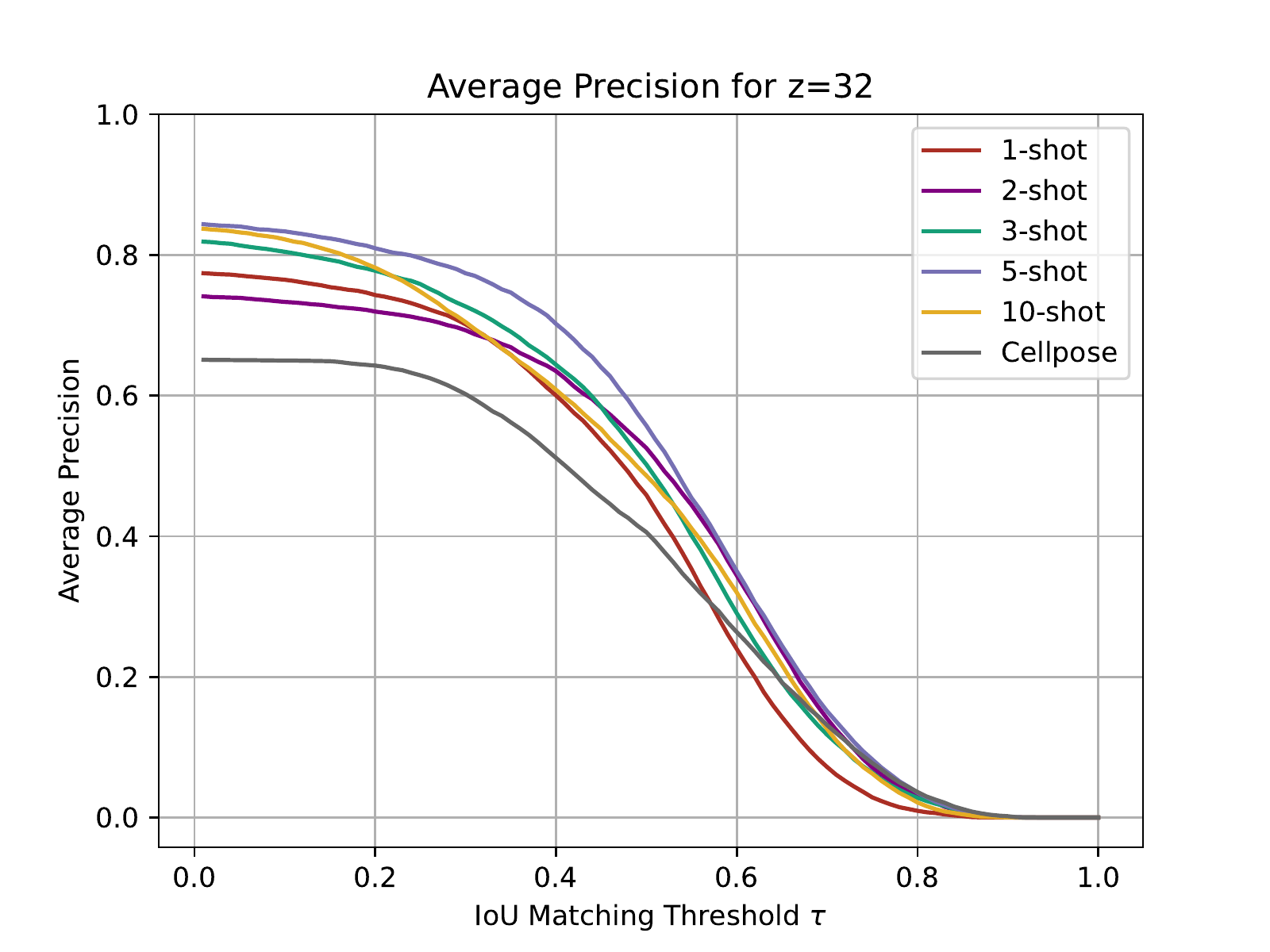}
\caption{\textbf{BBBC006 dataset.} Average precision of $K$-shot adaptation with $K= 1, 2, 3, 5, 10$ target samples. Experiments were completed using five BBBC006 datasets.}
\label{fig:AP}
\vspace{-4mm}
\end{figure*}

\noindent\textbf{Broad Bioimage Benchmark Collection -- 006.}
We evaluated CellTranspose on the target dataset BBBC006~\cite{Ljosa2012-si}, hosted by the Broad Institute. This dataset is composed of human U2OS cells which are fairly homogeneous and easy to segment in ideal settings. However, the same tissue samples have been imaged with different focus settings, generating different images, which allow us to observe the effect of the associated covariate shift on generalist and our approaches. We consider images taken with five focal planes, specifically at $\mathtt{z}$=00, 08, 16, 24, and 32. The optimal focal plane is at $\mathtt{z}$=16. Figure~\ref{fig:qualitative_results} shows qualitative segmentation results. As we move away from the optimal focal plane the generalist Cellpose model exposes greater performance deterioration than CellTranspose.

We also test CellTranspose with different numbers $K= 1, 2, 3, 5, 10$, of few-shot adaptation. Figure~\ref{fig:AP} plots the average precision (AP) against the intersection over union (IoU) for the different focal settings. CellTranspose consistently achieves high performance levels with as few as three annotated shots. Beyond this, results begin to exhibit diminished returns. So, unless otherwise specified, other results have been obtained with a 3-shot adaptation, given the balance it gives between performance and annotation needs. %requirement.

\noindent\textbf{TissueNet.}
We evaluated CellTranspose on TissueNet~\cite{Greenwald2021-uj}, a dataset developed alongside a generalist method called Mesmer. TissueNet is comprised of samples from various imaging platforms and tissue types, providing a wide-spanning array of cellular images. In~\cite{Greenwald2021-uj} one set of experiments split TissueNet into subsets of the four most common imaging types, each of which was further divided into different tissue types. Similarly, four other subsets were composed of the four most common tissue types, each being further split into the imaging types that make up the samples for that tissue type.
Table~\ref{table:tissuenet} shows the F-1 scores on these eight data splits, computed for the generalist Cellpose, Mesmer trained on each of the splits, and CellTranspose 3-shot adapted to each of the splits. Again, CellTranspose consistently shows improved performance. Since CellTranspose essentially adapts Cellpose to the target domain, we can interpret it as a \emph{lower bound} of performance. On the other hand, we also re-trained Cellpose with each target training dataset, obtaining what could be interpreted as an \emph{upper bound} for CellTranspose, and we indicated that as \textsl{Cellpose-UB}.

\begin{table}
\begin{center}
\caption{\textbf{TissueNet dataset.} F-1 score from state-of-the-art ``generalist'' approaches and CellTranspose.}
\label{table:tissuenet}
\begin{adjustbox}{width=\columnwidth}
\begin{tabular}{l|llll|llll}
\hline\noalign{\smallskip}
{\bf TissueNet} & \multicolumn{4}{c|}{\textbf{Platform-specific}} & \multicolumn{4}{c}{\textbf{Tissue-specific}} \\
{\bf Results} & CODEX & CyCIF & MIBI & Vectra & Breast & GI & Imm. & Panc.\\
\noalign{\smallskip}
\hline
\noalign{\smallskip}
Cellpose	& 0.785	& 0.548 & 0.479 & 0.609 & 0.670 & 0.523 & 0.350 & 0.797\\
Mesmer					& 0.88 	& 0.80	& 0.76	& 0.72	& 0.74	& 0.82	& 0.82	& 0.92\\
CellTranspose           & \textbf{0.940} & \textbf{0.940} & \textbf{0.932} & \textbf{0.918} & \textbf{0.911} & \textbf{0.906} & \textbf{0.934} & \textbf{0.955}\\
Cellpose-UB	& 0.962	& 0.967 & 0.945 & 0.960 & 0.950 & 0.940 & 0.953 & 0.958\\
\hline
\end{tabular}
\end{adjustbox}
\end{center}
\vspace{-8mm}
\end{table}

\noindent\textbf{Triple Negative Breast Cancer.}
Among the most challenging types of cellular data to segment is that of hematoxylin and eosin-stained (H\&E) images. This is in part due to the fact that multiple cell types often appear within an individual sample, in addition to the high variability of the background. The Triple Negative Breast Cancer (TNBC) dataset~\cite{Naylor2019-ak}, gathered by the Curie Institute, is comprised of 50 images obtained from 11 distinct tissue types, furthering the inherent difficulty of the dataset.
\setlength{\tabcolsep}{4pt}
\begin{table}[t]
\begin{center}
\caption{\textbf{TNBC dataset.} Comparison between the top unsupervised approach and CellTranspose. Best results are in bold, and second best results are underlined.}
\label{table:tnbc}
\begin{adjustbox}{width=\columnwidth}
\begin{tabular}{l|l|l|l}
\hline\noalign{\smallskip}
{\textbf{BBBC039 $\rightarrow$ TNBC}} & AJI & Pixel-F1 & Object-F1 \\
\noalign{\smallskip}
\hline
\noalign{\smallskip}
Cellpose									& 0.3815 $\pm$ 0.0794
												& 0.5829 $\pm$ 0.0689
												& 0.5408 $\pm$ 0.1124 \\

CyC-PDAM~\cite{Liu2020-qh}						& 0.5672 $\pm$ 0.0646
												& \textbf{0.7593 $\pm$ 0.0566}
												& 0.7478 $\pm$ 0.0417 \\
CellTranspose 3-shot 						& 0.4916 $\pm$ 0.0771
												& 0.6702 $\pm$ 0.0710
												& 0.7092 $\pm$ 0.0818 \\
CellTranspose 5-shot 						& \underline{0.5693 $\pm$ 0.0576} 
												& 0.7377 $\pm$ 0.0431
												& \underline{0.7825 $\pm$ 0.0625} \\
CellTranspose 10-shot						& \textbf{0.5906 $\pm$ 0.0617}
												& \underline{0.7568 $\pm$ 0.0493} 
												& \textbf{0.7879 $\pm$ 0.0687} \\
Cellpose-UB							& 0.5498 $\pm$ 0.0860
												& 0.7216 $\pm$ 0.0704
												& 0.7760 $\pm$ 0.0640 \\												
\hline
\end{tabular}
\end{adjustbox}
\end{center}
\vspace{-8mm}
\end{table}
\setlength{\tabcolsep}{1.4pt}

We compare CellTranspose with the top-performing unsupervised domain adaptive cellular instance segmentation approach, CyC-PDAM~\cite{Liu2020-qh}. This is the closest approach to ours we found in terms of problem settings and data, since the only other supervised domain adaptive instance segmentation approaches~\cite{Zhang2019-oj,Xu2018-om} were tuned to very specific and different applications. Following the lead of CyC-PDAM, we divert from the protocol used above and pretrain CellTranspose on the BBBC039 dataset. We then adapt the model with 3, 5, and 10 shot samples selected from the target dataset of 40 images and 8 tissue types from TNBC. Each model is then tested on the 10 images from the remaining 3 tissue types. Results, using the same metrics as CyC-PDAM, are shown in Table~\ref{table:tnbc}. It can be noted that a 5-shot adaptation leads to performance metrics comparable with those of CyC-PDAM, that has used all the training target data available to adapt the model in an unsupervised manner. Note that the settings of this experiment is still different and disadvantageous from the protocol for which CellTranspose was designed, because the target testing data distribution is different from the target training data distribution. We also tested Cellpose-UB in this settings, which appears to be affected more than CellTranspose by the domain shift still present in the evaluation protocol.
\begin{figure}[t]
\centering
\includegraphics[width=\columnwidth]{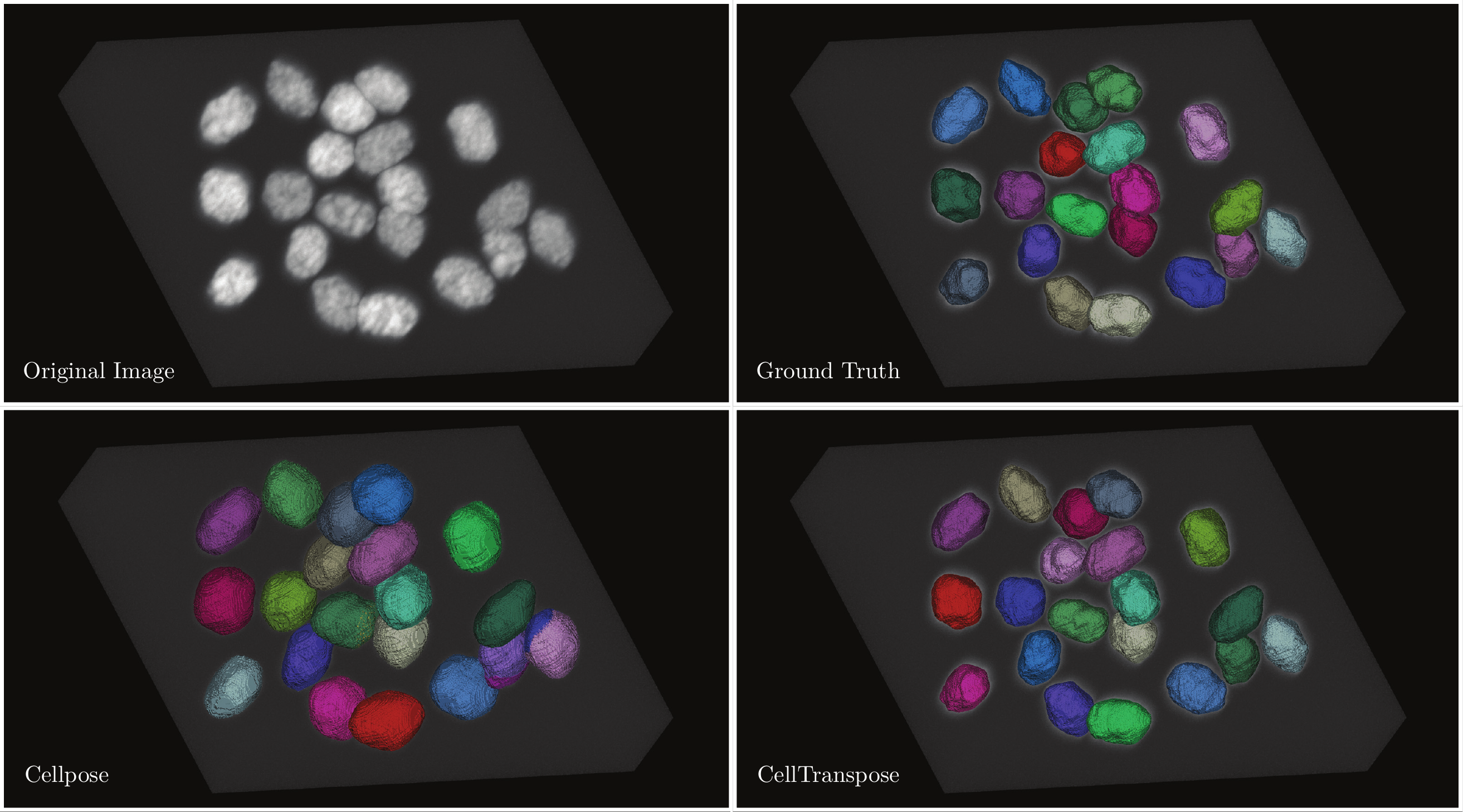}
\caption{\textbf{3D Segmentation.} Qualitative segmentation results on samples from the BBBC024 dataset. Note that Cellpose3D tends to oversample; this is likely due to the fact that the particular characteristics of the data Cellpose3D was trained on necessitated pixels of even lower intensities to be segmented. More importantly, on the bottom right of the results obtained with Cellpose3D, it is possible to observe some over-segmentation effects, which are not present in the results of CellTranspose3D.
}
\label{fig:3d_segmentation}
\end{figure}

\noindent\textbf{3D Segmentation.}
As biological applications often produce and require the analysis of 3-D data, following the approach in~\cite{Stringer2021-yw}, we extend CellTranspose to operate in 3-D, named \textsl{CellTranspose3D}, by making adaptation and combining the predictions along the $xy$, $yz$, and $zx$ volume sections.
We test our approach on two 3-D datasets. The BBBC024 is composed of synthetic annotated cells from the Broad Institute. The Worm dataset is a series of nuclei images from larval stage C. elegans. Similar to the 2-D experiments, only three 2-D sample patches, taken from the XY-plane from volumes in each training set, are used as target data for the model.
\begin{table}
\begin{center}
\caption{\textbf{3D datasets.} Average Precision at IoU threshold 0.5.}
\label{table:3D_results}
\begin{adjustbox}{width=0.5\columnwidth}
\begin{tabular}{l|l|l}
\hline\noalign{\smallskip}
{\bf 3-D Results} & {\bf Worm} & {\bf BBBC024}\\
\noalign{\smallskip}
\hline
\noalign{\smallskip}
Cellpose3D			    & 0.575 	& 0.822 \\
CellTranspose3D  			& 0.648 	& 0.994 \\
Cellpose3D-UB		    & 0.675 	& 1.0 \\
StarDist-3D				& 0.765 	& --- \\
\hline
\end{tabular}
\end{adjustbox}
\end{center}
\vspace{-5mm}
\end{table}

StarDist-3D~\cite{Weigert_2020_WACV} and Cellpose3D-UB have been trained and tested on Worm, and Cellpose3D-UB on BBBC024 also, and serve here as upper bounds. Table~\ref{table:3D_results} shows the average precision results, highlighting the improvement of CellTranspose3D over the generalist Cellpose3D. Figure~\ref{fig:3d_segmentation} provides a qualitative comparison between 3D segmentations of the BBBC024 dataset.

\noindent\textbf{Ablation Study.}
Table~\ref{table:ablation} shows ablation results computed on the BBBC006 dataset with $z$=00 and 3-shot adaptation. The addition of both contrastive losses improves the overall AP by more than 15\%. Interestingly, however, the removal of only one adaptation loss tends to decrease the performance to below that of removing both,  which follows the same training scheme outlined in Section~\ref{sec-adaptation}, but without either adaptation loss. This indicates that the flow and mask losses are intrinsically tied to one another, which is consistent with the fact that weights are shared between both outputs until the final layer.

Additionally, the cell size calculation method seems to play an important role in accurate segmentation. Cellpose computes the cell size based upon the number of pixels corresponding to a particular cell, but non-spherical cells have the potential to provide a similar diameter while spanning a much larger area. Thus, a more robust cell size calculation could be based on computing the total rectangular area enclosing a cell, which is how it is done in CellTranspose. The second row from bottom in Table~\ref{table:ablation} shows that when this second strategy is replaced with the one used by Cellpose, performance further deteriorates.

\begin{table}[t]
\begin{center}
\caption{\textbf{Ablation results on BBBC006.} AP is calculated for an IoU threshold of 0.5.}
\label{table:ablation}
\begin{adjustbox}{width=0.6\columnwidth}
\begin{tabular}{l|l}
\hline\noalign{\smallskip}
{\bf Ablation Results} & $AP_{50}$\\
\noalign{\smallskip}
\hline
\noalign{\smallskip}
CellTranspose & 0.509\\
No Contrastive Flow Loss & 0.414\\
No Contrastive Mask Loss & 0.434\\
No Adaptation Losses & 0.441\\
No Adaptation Losses \& No cell size & 0.390\\
Cellpose & 0.233\\
\hline
\end{tabular}
\end{adjustbox}
\end{center}
\vspace{-7mm}
\end{table}

%%% Local Variables:
%%% mode: latex
%%% TeX-master: "main"
%%% End:

\section{Conclusions}

In this work, we have exemplified the need for adaptation of cell instance segmentation methods when working in new domains. Our approach is effective at achieving this with very few samples, thanks to the proposed new losses and training procedures we implemented. CellTranspose shows a level of performance that effectively tackles the effects of the covariate shift, and that can be comparable to models fully retrained on the target distribution. We have also shown that few samples are sufficient to reach performance similar to unsupervised approaches that rely on larger datasets and training simulations. We found that 3 to 5 annotated samples usually allowed to match or surpass the state of the art. The adaptation procedure takes a few minutes on a single GPU, striking an attractive balance in terms of time and resources used for training and annotation.

\subsection*{Acknowledgements}

Research reported in this publication was supported by the National Institute Of Mental Health of the National Institutes of Health under Award Number R44MH125238. The content is solely the responsibility of the authors and does not necessarily represent the official views of the National Institutes of Health. This material is also based upon work supported by the National Science Foundation under Grant No. 1920920.

{\small
\bibliographystyle{ieee_fullname}
\bibliography{references}
}

\end{document}